\documentclass[letterpaper, 10 pt, conference]{ieeeconf}
\IEEEoverridecommandlockouts

\usepackage{cite}
\usepackage{amsmath,amssymb,amsfonts}
\usepackage{algorithmic}
\usepackage{graphicx}
\usepackage{textcomp}
\usepackage{multirow}
\usepackage{subcaption}
\usepackage{array}
\usepackage{xcolor}
\def\BibTeX{{\rm B\kern-.05em{\sc i\kern-.025em b}\kern-.08em
    T\kern-.1667em\lower.7ex\hbox{E}\kern-.125emX}}

\author{Alexey Kotcov, Maria Dronova, Vladislav Cheremnykh, Sausar Karaf and Dzmitry Tsetserukou%  
\thanks{The authors are with the Intelligent Space Robotics Laboratory, CDE, Skolkovo Institute of Science and Technology, Bolshoy Boulevard 30, bld. 1, 121205, Moscow, Russia.
{\tt \{alexey.kotcov, maria.dronova, vladislav.cheremnykh, sausar.karaf, d.tsetserukou \}@skoltech.ru}}
}
\title{AirNeRF: 3D Reconstruction of Human with Drone and NeRF for Future Communication Systems\\
}

\usepackage{caption}
\begin{document}
\maketitle
\begin{abstract}
In the rapidly evolving landscape of digital content creation, the demand for fast, convenient, and autonomous methods of crafting detailed 3D reconstructions of humans has grown significantly. Addressing this pressing need, our AirNeRF system presents an innovative pathway to the creation of a realistic 3D human avatar. Our approach leverages Neural Radiance Fields (NeRF) with an automated drone-based video capturing method. The acquired data provides a swift and precise way to create high-quality human body reconstructions following several stages of our system. The rigged mesh derived from our system proves to be an excellent foundation for free-view synthesis of dynamic humans, particularly well-suited for the immersive experiences within gaming and virtual reality. 
\end{abstract}

\section{Introduction}

Creating detailed 3D reconstructions of humans is a challenging topic in the research community, film, and game industries, where high-fidelity human recreation typically involves pre-recorded templates and multiple camera systems. However, such requirements extend beyond the application scenarios of general uses, such as personalised avatars for telepresence, AR/VR, Metaverse, virtual fitting, and more. Thus, the direct reconstruction of high-fidelity digital avatars from either video footage or image sets is of great practical importance. Classical methods, e.g. photogrammetry \cite{mikhail2001introduction}, have been used for the purpose of 3D object reconstruction, generating a dense point cloud of a scene from different perspectives. Nonetheless, it requires expensive equipment, a long setup time, and has limitations when it comes to measuring non-uniform surfaces and producing detailed reconstructions. Reflections and transparent objects can be challenging due to texture changes. Recently, the quality of 3D reconstruction has improved due to the advances in NeRF-based approaches. This paper, among other aspects, explores the potential applications of this technology in 3D human reconstruction for VR and telepresence purposes.

%However, establishing a convenient and high-fidelity solution using a lightweight capture setup remains a challenge. Previous solutions \cite{liu2021neural, 6468043} often required a dome-based multi-view setup for precise reconstruction and image-based rendering in novel perspectives.

In this study we present AirNeRF, a novel approach designed to create realistic personalized digital avatars for VR applications. It leverages advancements in data collection through drone-based imaging, novel NeRF-based view synthesis, 3D point cloud-driven mesh creation, segmentation, and automatic rigging.

\begin{figure}
    \centering
    \includegraphics[width=8.8cm]{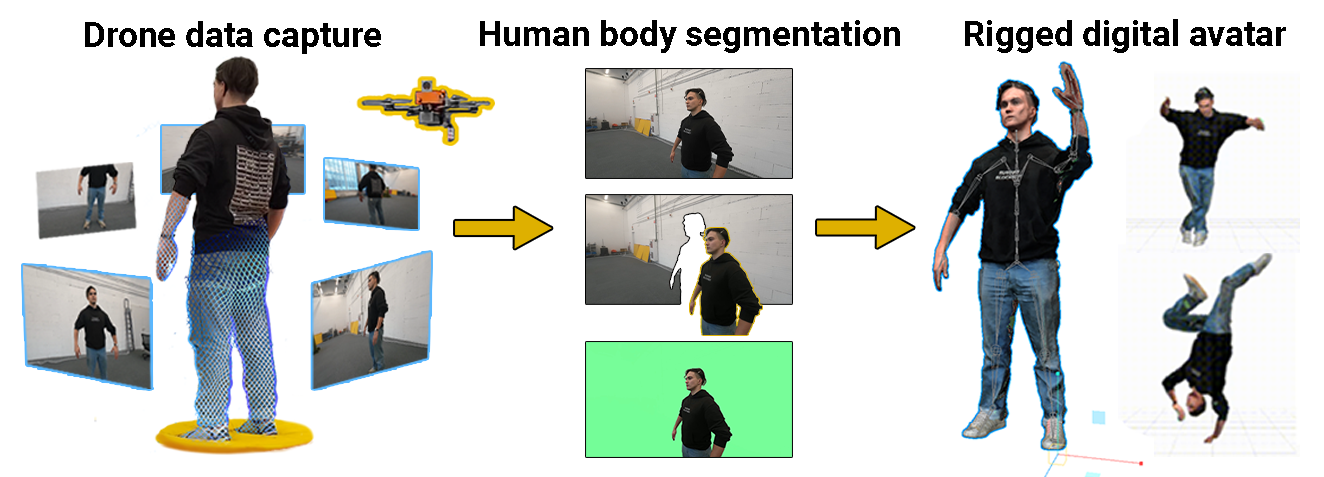}
    \caption{Layout of AirNeRF system for 3D human reconstruction.}
    \label{fig:system_overview}
\end{figure}

\section{Related Works}
Given that our research leverages NeRF as a backbone, we review existing literature related to NeRF, with a particular focus on realistic 3D human avatar creation. This focus is imperative for the rendering of human subjects within a scene, especially in the context of developing visual and immersive experiences \cite{liu2021neural, peng2023implicit}. We also briefly review works that aim to reconstruct and perform novel view synthesis of provided scenes.

\textbf{Neural Radiance Fields} have gained widespread popularity as a method for scene modeling and rendering from novel perspectives since their initial publication \cite{Tancik_2023}. This popularity stems from the high-quality rendering capabilities of NeRF \cite{Tancik_2023}. The inherent advantage of representing a scene as a radiance field lies in its built-in capability to render the scene seamlessly from various supported viewpoints through volume rendering.

Attempts have been made to customize NeRF for dynamic scenes \cite{park2021nerfies, park2021hypernerf, 10394246} and to manipulate and assemble scenes using different NeRF models \cite{guo2020objectcentric, dronova2024flynerf}, broadening their potential applications. While these approaches have produced intriguing and promising outcomes, they often necessitate the separate training of editable instances \cite{guo2020objectcentric} or manual curation of training data \cite{peng2023implicit}. However, this work focuses on a more feasible ‘in-the-wild’ setting. In the context of our specific task, considerable efforts have been directed towards NeRF models rigged modern methods, including neural networks such as RigNet \cite{RigNet} or SMPL \cite{10.1145/2816795.2818013} with the use of camera prediction algorithm COLMAP \cite{schoenberger2016sfm,schoenberger2016mvs}.

%Neural Body \cite{peng2023implicit}, for instance, assigns a latent code to each SMPL \cite{10.1145/2816795.2818013} vertex and utilizes sparse convolution to diffuse this latent code into the volume within the observation space.

\begin{figure*}[h!]
    \captionsetup[subfigure]{skip=0.5pt, justification=centering}
    \centering
    \begin{subfigure}[b]{0.19\textwidth}
        \centering
        \includegraphics[width=0.5\textwidth]{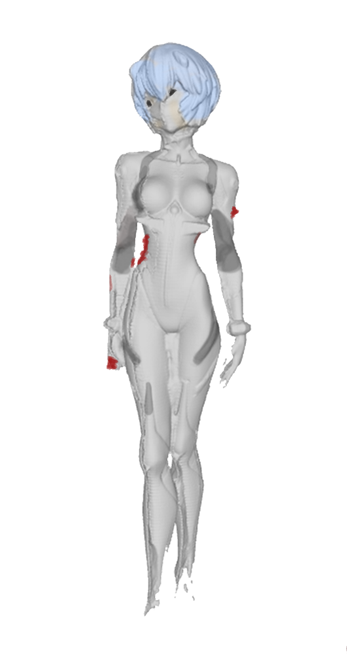}
        \caption{Photogrammetry\\(scanner)}
        \label{subfig:a}
    \end{subfigure}
    \begin{subfigure}[b]{0.19\textwidth}
        \centering
        \includegraphics[width=0.67\linewidth]{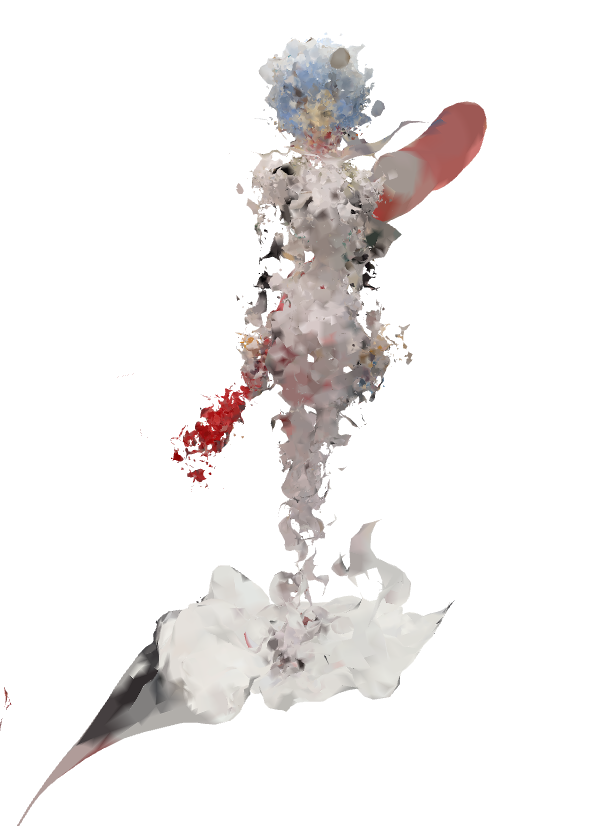}
        \caption{Photogrammetry\\(images)}
        \label{subfig:b}
    \end{subfigure}
    \begin{subfigure}[b]{0.19\textwidth}
        \centering
        \includegraphics[width=0.76\linewidth]{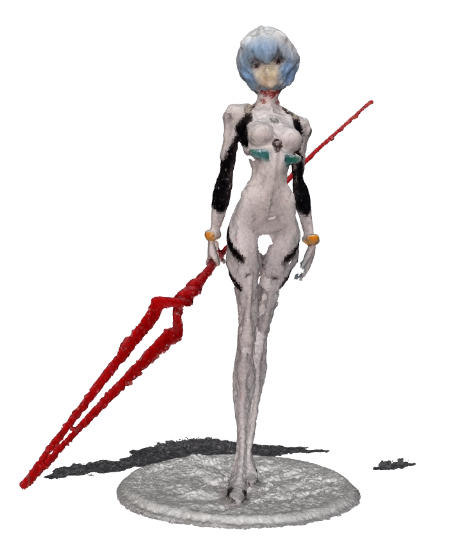}
        \caption{Nerfacto\\model}
        \label{subfig:c}
    \end{subfigure}
    \begin{subfigure}[b]{0.19\textwidth}
        \centering
        \includegraphics[width=0.8\linewidth]{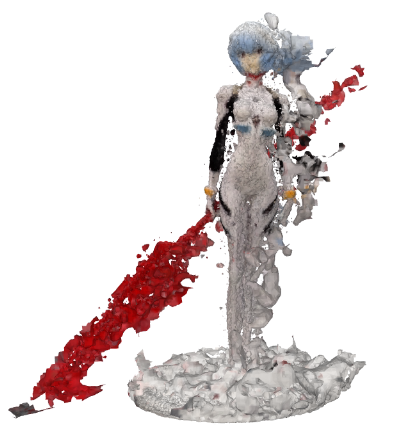}
        \caption{Instant-NGP\\model}
        \label{subfig:d}
    \end{subfigure}
    \begin{subfigure}[b]{0.19\textwidth}
        \centering
        \includegraphics[width=0.7\linewidth]{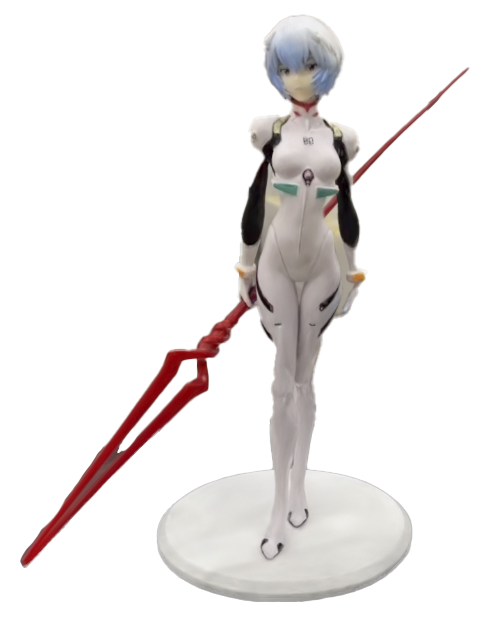}
        \caption{Gaussian Splatting\\render}
        \label{subfig:e}
    \end{subfigure}
    
    \caption{Comparison of 3D reconstruction methods: a) Photogrammetry using scanner. b) Photogrammetry from the set of images. c) Nerfacto model. d) Instant-NGP model. e) Gaussian Splatting render, on a static reduced-size human figure.}
    \label{fig:compar}
\end{figure*}

%Neural Actor \cite{liu2021neural} and NeuMan \cite{jiang2022neuman} employs volume warping based on SMPL \cite{10.1145/2816795.2818013} mesh transformation to learn human representations in canonical space. Additionally, it enhances rendering quality by incorporating a texture map. ST-NeRF \cite{zhang2021editable} divides the human into 3D bounding boxes, learning dynamic human elements within each box. While it does not necessitate precise human geometry estimation, it is limited in extrapolating to unseen poses due to its reliance on time-dependent frames. However, these methods all rely on an expensive multi-camera setup for acquiring ground truth information, making them unsuitable for our objective of reconstructing and rendering a human from a single video without additional devices or annotations and accounting for potential pose estimation errors.

%Neural Actor \cite{liu2021neural} and NeuMan \cite{jiang2022neuman} employs volume warping based on SMPL \cite{10.1145/2816795.2818013} mesh transformation to learn human representations in canonical space. While it does not necessitate precise human geometry estimation, it is limited in extrapolating to unseen poses due to its reliance on time-dependent frames. However, these methods all rely on an expensive multi-camera setup for acquiring ground truth information, making them unsuitable for our objective of reconstructing and rendering a human from a single video without additional devices or annotations and accounting for potential pose estimation errors.

Concurrently, HumanNeRF \cite{Zhao_2022_CVPR} aims for the free-viewpoint rendering of humans from a single video. While sharing similar goals with our approach, there are a few main distinctions. Firstly, HumanNeRF \cite{Zhao_2022_CVPR} depends on manual mask annotation for human-background separation, while our method utilizes modern segmentation techniques such as Semantic Guided Human Mating Segmentation \cite{chen2022sghm}. Secondly, HumanNeRF \cite{Zhao_2022_CVPR} represents motion using a combination of skeletal and non-rigid transformations, potentially resulting in ambiguous or unknown transformations under novel poses. In contrast, our method addresses this ambiguity by utilizing auto rigging methods, which allows to project joints on a 3D model of the human body. A similar study, Vid2Actor \cite{weng2020vid2actor}, constructs animatable humans from a single video by jointly learning a voxelized canonical volume and skinning weights. Despite similar objectives, our method achieves precise human geometry reconstruction using less than 300 images, compared to Vid2Actor's requirement for thousands of frames. This demonstrates the data-efficiency of our approach.

\textbf{Human body reconstruction.} This approach has seen widespread adoption of markerless techniques for achieving free-viewpoint videos or reconstructing geometric representations. Some recent endeavors have embraced a lightweight and single-view setup \cite{ 10.1145/3311970, 9157340}. However, these methods necessitate pre-scanned templates or naked human models, presenting challenges in achieving photo-realistic view synthesis. On the other end of the spectrum, high-end approaches \cite{habermann2021, liu2021neural, 6468043, 6126338} demonstrate the capability to deliver high-quality surface motion and appearance reconstruction. Nonetheless, they demand dense camera setups and a controlled imaging environment, posing accessibility challenges.

The SelfRecon method was proposed in the following work \cite{jiang2022selfrecon}. It merges implicit and explicit representations to achieve coherent space-time reconstructions of clothed human bodies from monocular self-rotating video input. As a result, it gives a high-fidelity deformable 3D human body reconstruction, though it has some limitations. Its optimisation process is time-consuming, which limits its practical use. Moreover, relying on a normal prediction network may result in inaccurate predictions for real-world applications and inconsistency across different frames. 

Monocular RGB-D-based methods \cite{8578859, 10.1145/3072959.3083722, jiang2022neuralhofusion, 8778689,lightairisr}  adopt traditional modeling and rendering pipelines for synthesizing novel human views. Despite their contributions, these methods grapple with the inherent self-occlusion constraints, limiting their ability to capture motions in occluded regions. Lightweight multi-view solutions \cite{10.1145/3130800.3130801, 10.1145/2897824.2925969, 8708933}, resembling our approach the most, strike a balance between hardware demands and high-fidelity reconstruction. However, these solutions still rely on 3 to 8 RGB-D streams as input.

\section{Quality Comparison of the 3D Reconstruction Approaches} \label{sec:compar}

The first stage of our study involved selecting the suitable method for 3D reconstruction. The primary criteria we chose dictate that a method must be both fast and efficient, enabling seamless integration into the AirNeRF pipeline for human body avatar creation. We performed the first comparison step on a static model of a reduced-size human figure, thereby eliminating the dynamic factors introduced when working with a person. In the second step, a comparison was conducted using a human model.

\subsection{Static model}

We performed a comparative analysis that included conventional photogrammetry approaches and novel volume rendering techniques, such as NeRF-based and Gaussian Splatting methodologies. The traditional method, in turn, comprised two approaches: the first included the use of an Artec 4 scanner and Artec Studio software, and the second entailed reconstructing the object from a set of images by the Agisoft Metashape software. The results of the comparison are illustrated in Table~\ref{tab:method_comp_stat} and Fig.~\ref{fig:compar}.

\begin{table}[t]
    \centering
    \caption{Comparison of Methods on Static 3D Model}
    \begin{tabular}{|m{2cm}|>{\centering\arraybackslash}m{0.7cm}|c|c|>{\centering\arraybackslash}m{1.05cm}|}
        \hline
        & \textbf{Time, min} & \textbf{Vertices}  & \textbf{Triangles} & \textbf{File size, Mb} \\
        \hline
        \textbf{Photogrammetry (scanner)} & 30 & 648 861 & 1 297 650 & 23.5 \\
        \hline
        \textbf{Photogrammetry (images)} & 16 & 73  319 & 143 026 & 11 \\
        \hline
        \textbf{NeRF \newline(Nerfacto)} & 6 & 230 749 & 455 547 & 17.7 \\
        \hline
        \textbf{NeRF \newline(Instant-NGP)} & 5 & 207 362 & 402 727 & 15.8 \\
        \hline
        \textbf{Gaussian \newline splatting} & 5 & 557 002 & 0 & 138 \\
        \hline
    \end{tabular}
    \label{tab:method_comp_stat}
\end{table}

%The process of obtaining a reconstruction using Artec 4 works as follows: first, the scanner is connected to the computer and configured. The model for reconstruction is subsequently maneuvered at a steady pace, maintaining a distance of 10 cm from its surface, while the scanner is held perpendicular to it. Model tracks containing information about its surface are then combined in the Artec Studio. After complete merging, it is necessary to post-process the mesh to remove excess vertices and smooth the surface. 

%Reconstruction using Agisoft Metashape application requires a set of images capturing the model from multiple angles along a circular trajectory. Subsequently, the photos are spatially combined to produce a point cloud. Obtaining a mesh from point cloud involves several steps: normals estimation with a tangent plane limitation, Poisson surface reconstruction and further mesh optimization, which includes removing unreferenced vertices, degenerate triangles and non-manifold edges.

When comparing the latest methods with traditional ones, we have selected the following examples: the Nerfacto model, which comprises elements of diverse NeRF-based methodologies for efficient volume rendering, the Instant-NGP model, distinguished by its enhanced sampling technique compared to vanilla NeRF, and Gaussian Splatting, a novel approach for representing radiance fields by explicitly storing a collection of 3D volumetric Gaussians.

All of the above methods require a dataset consisting of images of the model and the corresponding spatial positions of the camera. In this study, we acquired these datasets by utilizing a video capture of the model and the COLMAP application \cite{schoenberger2016sfm,schoenberger2016mvs}. To ensure a fair comparison of the methods, the images used for training Gaussian Splatting-based and Gaussian Splatting models, as well as photogrammetry with images, were identical.

Resulting Table~\ref{tab:method_comp_stat} presents key metrics crucial for our application, including the time required to generate a mesh from having a physical model, vertices and triangles of the output mesh, and file size of the PLY format in MB. Upon analyzing these results, we have concluded that neither of the photogrammetry methods fulfills our requirements in terms of small processing time and the number of triangles in the model. Overload in triangles number may result in a lack of engine support, while an insufficient amount may lead to non-representative avatar model. It is noteworthy that Gaussian Splatting demonstrated excellent performance according to render visualization results, as can be seen in Fig.~\ref{fig:compar}(e). However, it should be taken into account that due to its methodological specifics, Gaussian Splatting has its own model representation that differs from conventional triangles or polygons. This representation needs further post-processing to achieve a standard 3D model format. Such a pipeline requires high-power GPU efficiency and sufficient preprocessing time. The aforementioned factors compel us to discard this method in favor of developing an effective and swift system for reconstruction.

\begin{table*}
    \centering
    \caption{Comparison of Methods on 3D Human Model}
    \begin{tabular}{|c|c|c|c|c|c|c|c|}
        \hline
        & \textbf{Time, min} & \textbf{Vertices}  & \textbf{Triangles} & \textbf{PSNR} & \textbf{SSIM} & \textbf{LPIPS} & \textbf{File size, Mb} \\
        \hline
        \textbf{NeRF \newline(Nerfacto)} & 7 & 209 288 & 413 858 & 19.761 & 0.921 & 0.117 & 15.3 \\
        \hline
        \textbf{NeRF  \newline (Instant-NGP)} & 6 & 641 241 & 1 234 369 & 19.238 & 0.915  & 0.129  & 46.4 \\
        \hline
    \end{tabular}
    \label{tab:method_comparison}
\end{table*}

\subsection{Human model}

During this experiment, photogrammetry with a scanner encountered challenges when reconstructing large objects such as the human body. This particular case necessitates the use of larger and more expensive scanners. Additionally, as we concluded from the last experiment, the obtained results are user-dependant due to the manual nature of the scanning process. Furthermore, achieving a comprehensive 3D reconstruction of the human body through this method demanded extensive post-processing manipulation within the Artec Studio software. The reconstruction of the human body from images using photogrammetry provides poor results as well. Based on this, as well as on the outcomes of the static model experiment, in this section we compare only the two most promising NeRF-based approaches yielding optimal results. 

Table~\ref{tab:method_comparison} presents a comparison of meshes obtained for a human model through the implementation of Nerfacto and Instant-NGP NeRF models. Given that the reconstruction quality metrics (PSNR, SSIM, and LPIPS) are nearly identical, we should favor the NeRF model that produces results with fewer triangles and vertices, as our main goal is to achieve suitable 3D model format for AR/VR engines support while ensuring high-quality 3D representation of the person. Based on the experimental results, we can conclude that the Nerfacto model performs better on application for our specific task of human body 3D reconstruction.

\section{System Pipeline}

In the following section, we provide a comprehensive description of the proposed AirNeRF system configuration for obtaining a human avatar. To create a digital avatar using our method, it is required to have a GPU capable of NeRF training, a drone, and an open space of at least 2 by 2 meters. Through this pipeline, our work aims to address the need for personalized avatars, potentially enhancing interaction within VR environments. 

\subsection{Drone configuration}

Fig.~\ref{fig:system_overview} provides an overview of the proposed system, which utilizes a single unmanned aerial vehicle (UAV) equipped with a DJI Action 2 camera to capture visual data for the construction of a realistic 3D human avatar. To ensure accurate 3D avatar reconstruction and autonomous scanning via the drone, a precise localization technique is necessary. While onboard sensors could potentially be used, this study employs the Vicon V5 tracking system for improved accuracy. The motion capture system consists of 14 infrared cameras detecting retroreflective markers placed on the UAV. By using the information about the markers' position, the tracking system can triangulate the UAV's position with high precision. The Vicon tracking system enables the UAV's location to be determined at a high update rate, allowing for safe and collision-free camera capture while maintaining a one-meter distance from the person. In our system, the refresh rate of the tracking system is set to 100 Hz.
\begin{figure}[h!]
    \centering
    \includegraphics[width=\linewidth]{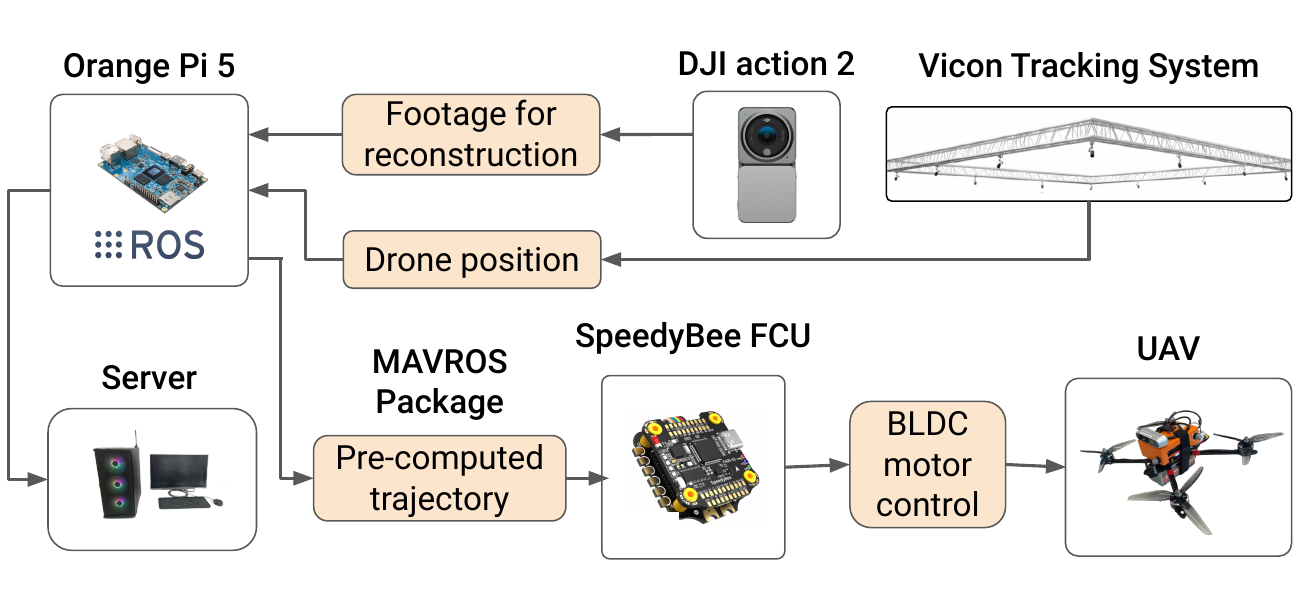}
    \caption{Hardware architecture of the AirNeRF system.}
    \label{fig:system_overview}
\end{figure}

The UAV setup includes an OrangePi 5 single-board computer, working in tandem with a SpeedyBee F405 flight controller and electronic speed controller stack. The single-board computer utilizes the Robot Operating System (ROS) framework for communications with the flight controller via the MAVROS package. Upon system initialization, a helical trajectory is calculated using the following equations:
\begin{equation}
\label{eq:helix_x}
x(\theta) = r cos(\theta) 
\end{equation}
\begin{equation}
\label{eq:helix_y}
y(\theta) = rsin(\theta) 
\end{equation}
\begin{equation}
\label{eq:helix_z}
z = \frac{h_{max} - h_{min}}{\theta_{max}}\theta - h_{min}
\end{equation}
where $r$ represents the radius of the helical shape, $h_{min}$ and $h_{max}$ are the minimum and maximum flight heights, respectively, and, $\theta_{max}$  is the maximum value of the parameter $\theta$. As the value of $\theta$ increases the points $(x(\theta), y(\theta), z(\theta))$ maintain a helical trajectory. The drone then flies through all points of the trajectory while maintaining the camera's orientation towards the center of the helix. Once the scanning process is complete, the drone returns to its starting position and lands.

\subsection{NeRF model configuration}
As an integral component of our system, the choice of NeRF model configuration plays a crucial role in achieving optimal results, as we discussed in Section~\ref{sec:compar}. Among the spectrum of different NeRF models, the Nerfacto emerged as our preferred selection due to its fast processing time and high quality of resulting reconstruction. Nerfacto is under active development and incorporates the advantages of various approaches of model implementation using different parts from recent researches that are proven to be effective. This model's distinctive ability to handle complex spatial patterns aligns seamlessly with our objective of crafting detailed 3D reconstructions of humans.

During the training phase, our model underwent 30,000 iterations, a process driven by a dataset comprising 300 RGB images collected by the drone. This approach ensures the robust learning of intricate features necessary for the accurate representation of human subjects. 

Executing on the Ubuntu 22.04 system, powered by the RTX 4090 graphics card, the Nerfacto configuration provided the ability for efficient and precise digital content creation. This strategic selection formed the backbone of our system and enabled to achieve the rapid and autonomous generation of high-quality 3D reconstructions with the training process completed in 6 minutes on average. 

\section{System Evaluation} \label{sys_eval}

We now provide details on the method used in our experiments for reconstructing digital human avatars. Our overall architecture consists of several stages, which we describe in the following subsections.

\subsection{Dataset collection}

\begin{figure}[h!]
    \centering
    \includegraphics[width=5cm]{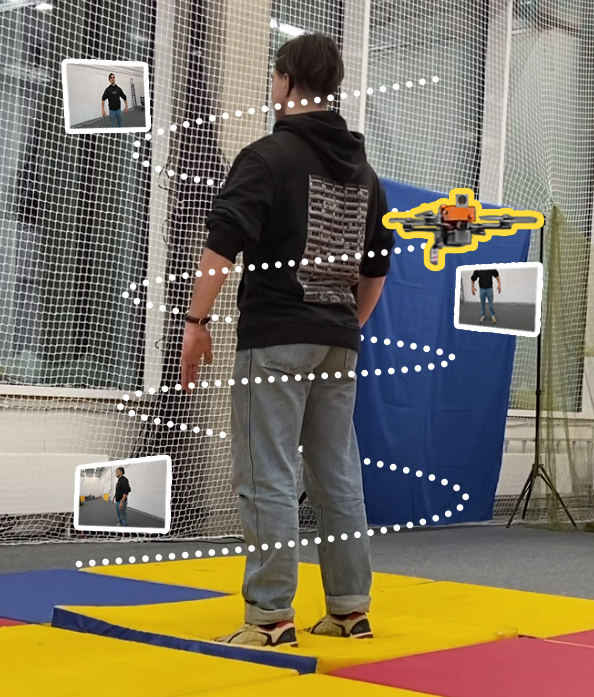}
    \caption{Data capturing process by the drone following the helical trajectory.}
    \label{fig:helix-drone}
\end{figure}
Our data collection approach aims to be autonomous and robust, as this stage directly impacts the quality of the final reconstruction. We use a drone rather than manual image capture to reduce human error and variability during acquisition. The drone allows us to precisely program reproducible scanning trajectories around the subject. For this work, we implement a helical trajectory maintaining a constant distance of one meter to the scanned human while capturing a full 360-degree view from bottom to top, as depicted in Fig.~\ref{fig:helix-drone}. This helical path provides stability throughout the footage and maximizes the information captured about the human body from all angles. 

Avoiding suboptimal viewpoints and occlusion is crucial for generating complete and accurate reconstructions. By using an automated drone-based capture method, we obtain consistent high-quality data that fully observe the subject. This improves reconstruction fidelity compared to casual handheld footage. Our experiments demonstrate the benefits of this controlled and robust data collection approach before segmentation, NeRF training, and rigging stages which rely on comprehensive input.

\subsection{Human body segmentation} 

\begin{figure} [h!]
    \centering
    \includegraphics[width=8cm]{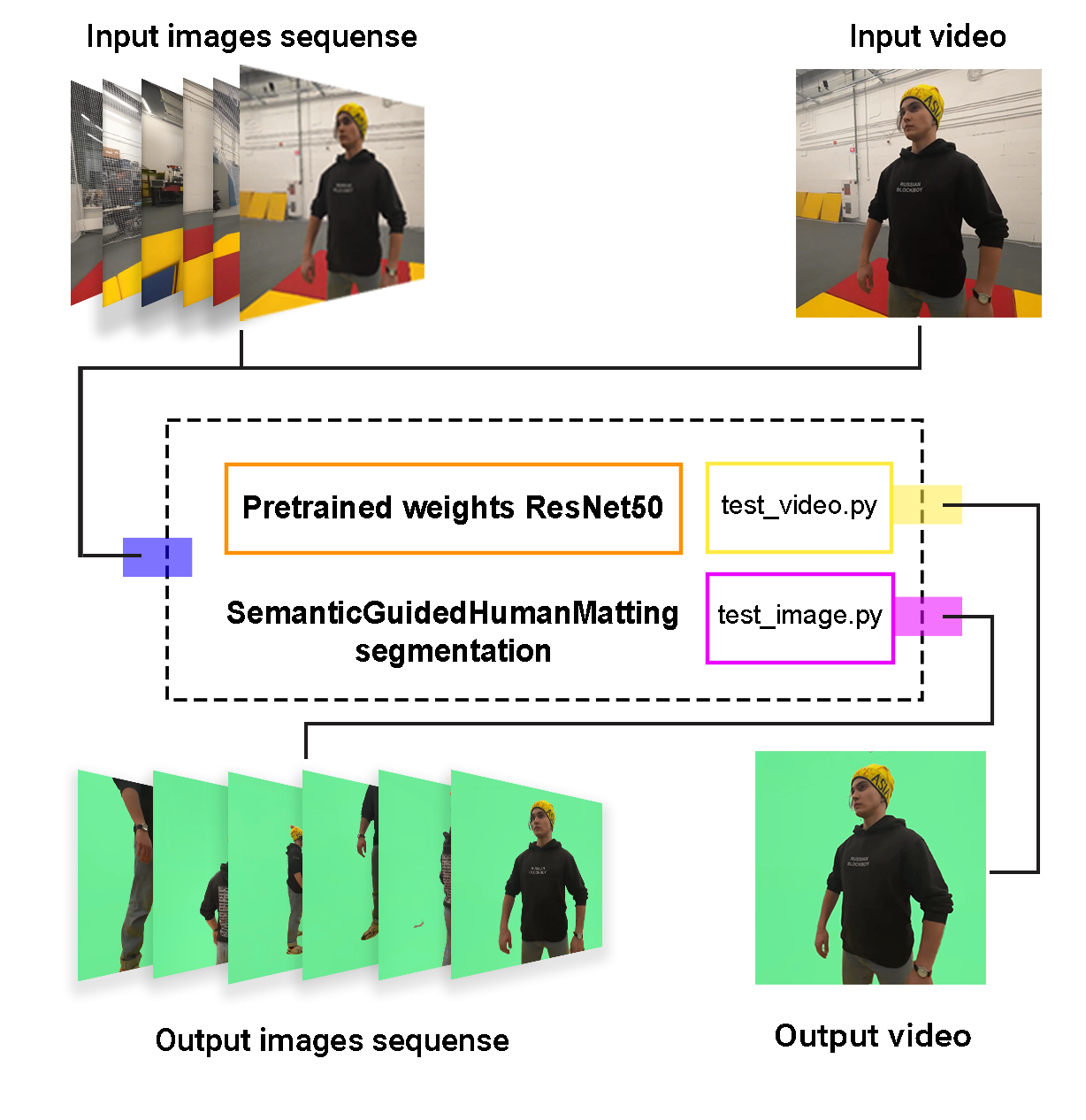}
    \caption{Segmentation pipeline for extracting human outline from the collected data.}
    \label{fig:segmentation}
\end{figure}

After obtaining the multi-view dataset, preprocessing and segmentation of the human subject are necessary preparatory steps before NeRF reconstruction. Segmenting out the human body from each frame helps focus the subsequent reconstruction on the relevant subject while ignoring unnecessary computation on background regions, therefore speeding up the overall NeRF training process. The segmentation pipeline used in our system is shown in Fig. \ref{fig:segmentation}. 

We implement human matting using a state-of-the-art SemanticGuidedHumanMatting module with a pretrained ResNet50 backbone \cite{he2016deep}. This network processes each input frame to extract pixel-wise human segments based on semantic guidance. Depending on the inputs, the module outputs either a folder of processed images or a segmented video sequence. By removing background clutter, the segmentation preprocesses the data into a domain-specific set focused on reconstructing the human. The resulting segments can then be passed directly to the NeRF training stage, avoiding excess computation and accelerating avatar reconstruction. In summary, targeted preprocessing transforms the raw captured footage into a concise representation adapted for full-body 3D human reconstruction from multiple views.

\subsection{NeRF training}

\begin{figure} [h!]
    \centering
    \includegraphics[width=8cm]{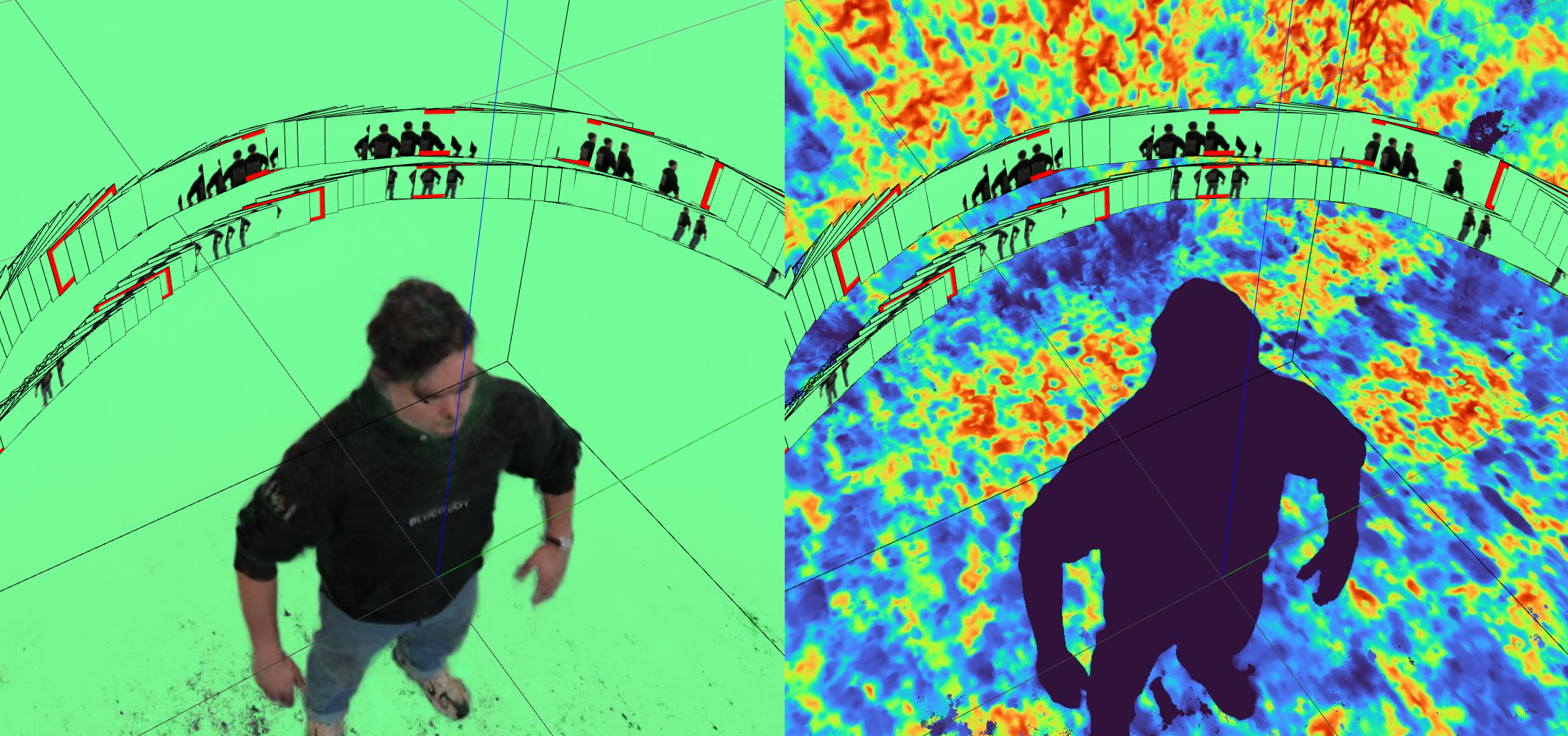}
    \caption{NeRF training process.}
    \label{fig:nerf-train}
\end{figure}

At this stage, we have fully prepared data for training the NeRF model, either as an image sequence or as a video file. Prior to starting the training process, it is necessary to compute the positional data for each image within the global coordinate system. To accomplish this task, we employ the COLMAP Structure from Motion (SfM) tool. In our experiment, we input a video in MOV format, necessitating the configuration of COLMAP to use sequential matcher and SIFT features. These parameters are more appropriate for detecting poses from video data. Following the completion of the positional calculations, the training phase is started. The visual representation in Fig.~\ref{fig:nerf-train}, with the depth channel displayed on the right side, illustrates the efficacy of our segmentation preprocessing. Notably, only one object is reconstructed within the scene, according to the intended outcome of the process.

\subsection{Mesh rigging}

\begin{figure}
    \centering
    \begin{subfigure}[b]{0.4\textwidth}
        \centering
        \includegraphics[width=\textwidth]{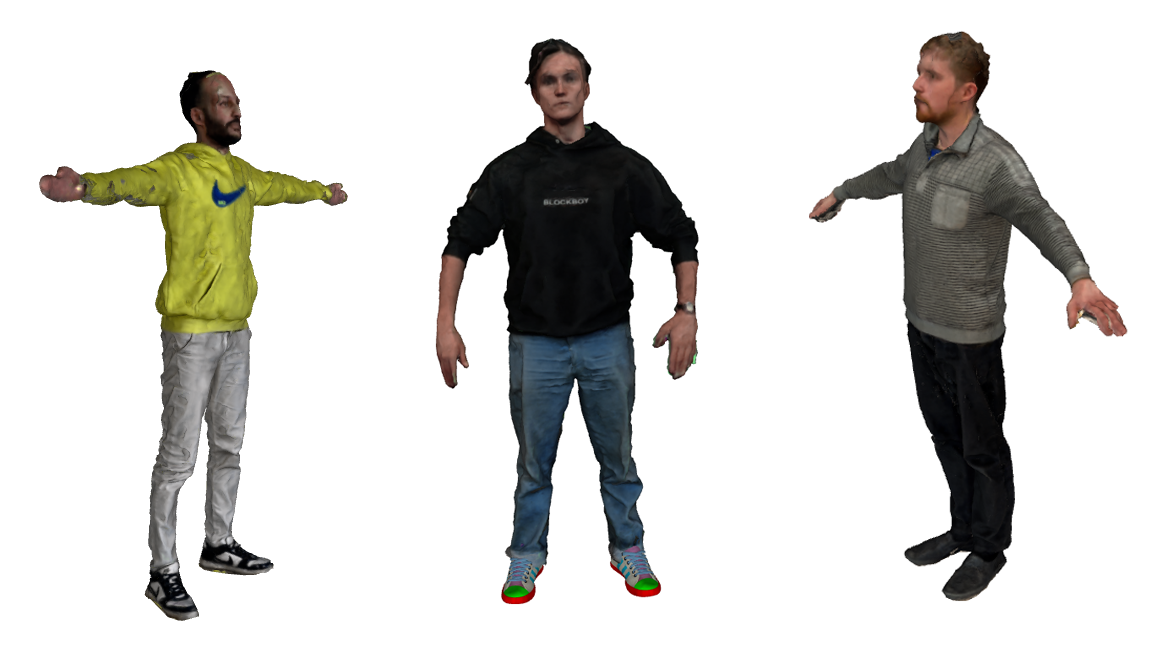}
        \caption{Human avatar models}
        \label{fig:subfiga}
    \end{subfigure}
    \hfill
    \begin{subfigure}[b]{0.4\textwidth}
        \centering
        \includegraphics[width=\textwidth]{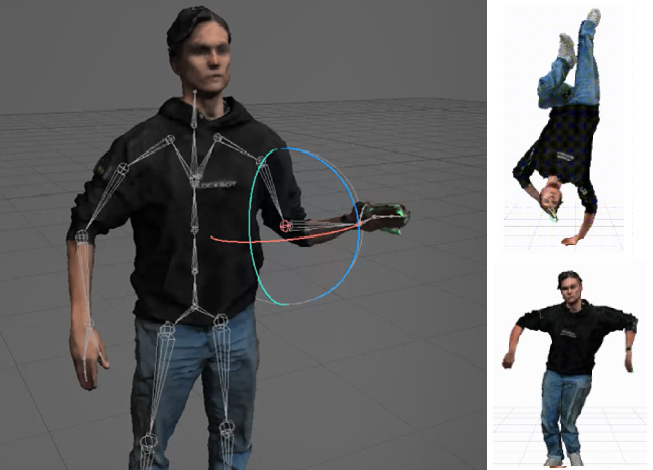}
        \caption{Rigged digital avatar after AccuRIG processing}
        \label{fig:subfigb}
    \end{subfigure}
    \caption{System results: digital avatars for rigging.}
    \label{fig:avatars}
\end{figure}

During the dataset collection phase, our volunteers stand in a specific A-pose or T-pose (Fig.~\ref{fig:avatars}(a)) that is required for the auto-rigging tool AccuRIG. The tool is used to estimate the correct joint positions of the human body with precision. After processing our static human mesh through the auto-rigging module, the output comprises a rigged human body including 22 joints. This amount of joints, particularly the additional two for each hand and each leg, enables the model to make natural rotations and movements. As a result, we have a skeleton that meets the requirements for applying any modern humanoid animation file to the final digital avatar, as can be seen in Fig.~\ref{fig:avatars}(b). The Table~\ref{tab:comparison} presents surface characteristics of the resulting rigged avatars, as well as the file size of the FBX format. 

\begin{table}[h!]
\centering
\caption{Reconstructed 3D Models}
\begin{tabular}{|c|c|c|>{\centering\arraybackslash}m{0.6cm}|>{\centering\arraybackslash}m{2,5cm}|}
\hline
& \textbf{Vertices} & \textbf{Triangles} & \textbf{Size, Mb} & \textbf{Supported Engines} \\
\cline{1-5}
\textbf{Human 1} & 150756 & 50252 & 2.81  & \multirow{3}{*}{
\begin{tabular}
[c]{@{}c@{}}Unity,  Unreal \\  Engine, Lens Studio,\\ Meta Spark Studio
\end{tabular}} \\
\cline{1-4}
\textbf{Human 2} & 108594 & 36198 & 2.71 & \\
\cline{1-4}
\textbf{Human 3} & 154254 & 51418 & 3.00 & \\
\hline
\end{tabular}
\label{tab:comparison}
\end{table}

As indicated by the metrics, amount of triangles ranges from approximately 36,000 to 51,000 which meets modern 3D engines requirements for triangle count. The resulting file size for all avatars falls within a single range and is relatively lightweight. The following results allow the utilization of the acquired avatars in various popular game engines. 

\section{Conclusion and Future Work}
We have developed AirNeRF, a novel system for generating the digital human avatars by leveraging NeRF and drone-based motion capturing. The proposed pipeline features NeRF rendering and automatic rigging to digitize a real human without 3D modeling.

The results of the research experiment, as demonstrated by the rigged avatars and their corresponding metrics, showcase the system's capability to generate fully prepared human avatars for 3D engines from lightweight drone footage. Specifically, the resulting avatars comprise 22 joints, with a triangle count ranging from approximately 36,000 to 51,000, meeting modern 3D engine requirements. The file sizes of the avatars, in FBX format, fall within a lightweight range, around 3 MB, making them easily integrated into various popular game engines.

VR, AR, and Metaverse fields can be enhanced in terms of user experience by lowering the threshold for the human avatar creation stage preserving realistic appearance and correct proportions. AirNeRF system can potentially considerably improve the immersion of the human into personalized Metaverse and 3D teleconferencing, making such technologies affordable for everyone. Additionally, Industry 4.0 and BIM can potentially benefit from 3D reconstruction of the large scale objects, e.g. engines and buildings. In further research, our development group will focus on the possibility of enhancing the system through the implementation of a Gaussian Splatting mesh reconstruction pipeline.

\section*{Acknowledgements} 
Research reported in this publication was financially supported by the Russian Science Foundation grant No. 24-41-02039.

\bibliographystyle{ieeetr}
\bibliography{sample-base}

\end{document}